\documentclass[journal]{IEEEtran}
\usepackage{cite}
\usepackage{booktabs}
\usepackage{verbatim}
\usepackage{amsmath}
\usepackage{subfigure}
\usepackage{amssymb}
\usepackage{color}
\usepackage{bbding}
\usepackage{diagbox}
\usepackage{url}
\ifCLASSINFOpdf
  \usepackage[pdftex]{graphicx}
\else
\fi
\begin{document}

%
\title{SSPNet: Scale Selection Pyramid Network for Tiny Person Detection from UAV Images}
%

\author{Mingbo~Hong,
        Shuiwang~Li,
        Yuchao~Yang,
        Feiyu~Zhu,
        Qijun~Zhao,
        Li~Lu
\thanks{Mingbo Hong, Shuiwang Li,Yuchao Yang, Feiyu Zhu, Qijun Zhao and Li Lu are with College of Computer Science, Sichuan University, China, 30332 USA e-mail: (kris@stu.scu.edu.cn, lishuiwang0721@163.com, yangyuchao@stu.scu.edu.cn, feiyuz@scu.edu.cn, qjzhao@scu.edu.cn, luli@scu.edu.cn)}
}

%
%

\markboth{Manuscript}%
{Shell \MakeLowercase{\textit{et al.}}: Bare Demo of IEEEtran.cls for IEEE Journals}
%



\maketitle

\begin{abstract}
With the increasing demand for search and rescue, it is highly demanded to detect objects of interest in large-scale images captured by Unmanned Aerial Vehicles (UAVs), which is quite challenging due to extremely small scales of objects. 
Most existing methods employed Feature Pyramid Network (FPN) to enrich shallow layers' features by combing deep layers' contextual features. However, under the limitation of the inconsistency in gradient computation across different layers, the shallow layers in FPN are not fully exploited to detect tiny objects.
In this paper, we propose a Scale Selection Pyramid network (SSPNet) for tiny person detection, which consists of three components: Context Attention Module (CAM), Scale Enhancement Module (SEM), and Scale Selection Module (SSM). 
CAM takes account of context information to produce hierarchical attention heatmaps. SEM highlights features of specific scales at different layers, leading the detector to focus on objects of specific scales instead of vast backgrounds.
SSM exploits adjacent layers' relationships to fulfill suitable feature sharing between deep layers and shallow layers, thereby avoiding the inconsistency in gradient computation across different layers.
Besides, we propose a Weighted Negative Sampling (WNS) strategy to guide the detector to select more representative samples.
Experiments on the TinyPerson benchmark show that our method outperforms other state-of-the-art (SOTA) detectors.
\end{abstract}


%
\IEEEpeerreviewmaketitle

\section{Introduction}
%
%
%
%

\IEEEPARstart{A}{s} a high-efficiency image acquisition system, UAVs have the advantages of high intelligence, high mobility and large field-of-view, and have thus been widely used in the emerging field for searching persons in a large area and at a very long distance. However, in such a scenario, finding persons is challenging since most persons in the obtained images are of tiny scale with low signal-noise ratio and easily contaminated by backgrounds\cite{yu2020scale}.

\begin{figure}[htbp]
	\centering
	\includegraphics[width=9cm,height=3.8cm]{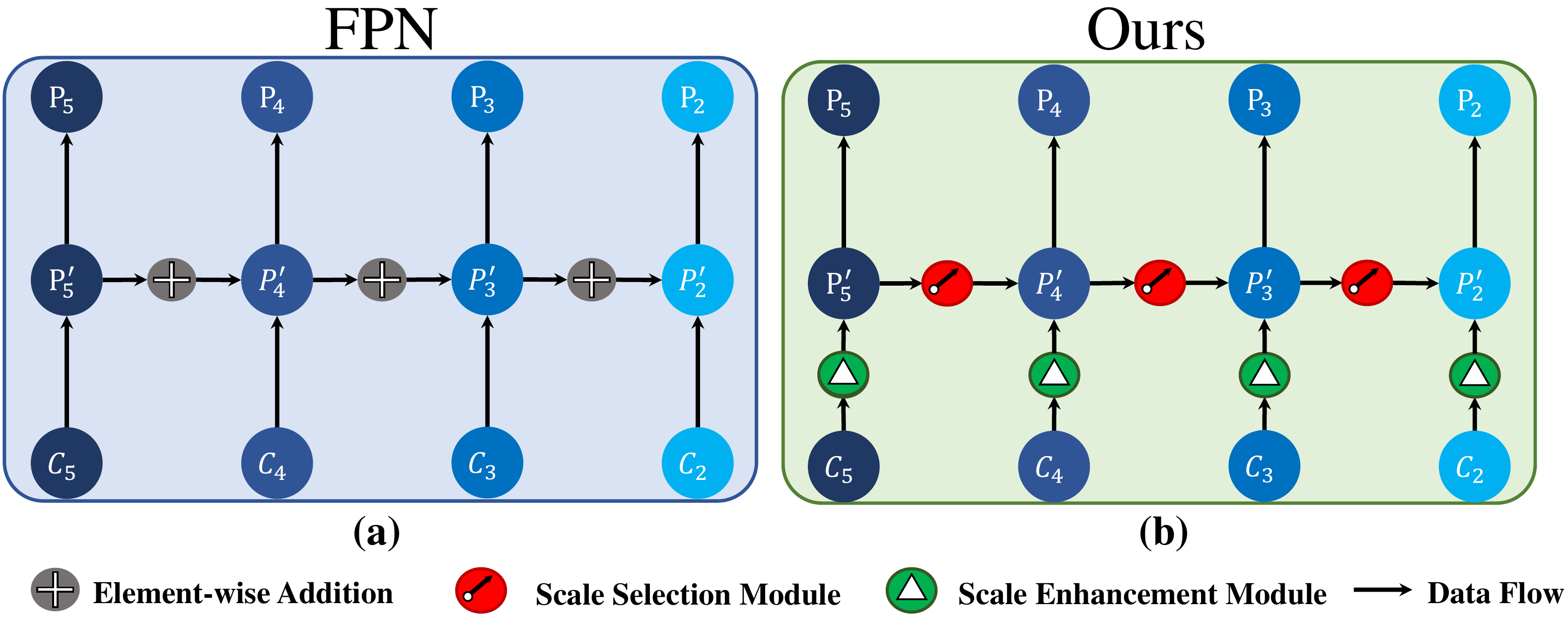}
	\caption{Illustrations of FPN (a) and our SSPNet (b). FPN employs the element-wise addition operation to integrate the adjacent features directly. Our SSPNet can learn adjacent layers' relationships to fulfill suitable feature sharing between deep layers and shallow layers, thereby avoiding the inconsistency in gradient computation across different layers.}
	\label{fig:challenge}
\end{figure}

A couple of methods have recently been proposed for tiny object detection. Some of them employ multi-training stages to enrich features to improve performance\cite{bai2018finding,noh2019better}, and others use a general-purpose framework with data augmentation to improve the detection of tiny objects\cite{kisantal2019augmentation,yu2020scale,yu20201st}.
Since deeper layers may furnish more semantic features, most existing methods\cite{liu2019learning,kong2020foveabox} employed  Feature Pyramid Network (FPN) to enrich shallow layers’ features for boosting performance by integrating deep layers’ features.
Gong et al.\cite{Gong_2021_WACV} propose a feature-level statistic-based approach to avoid propagating too much noise from deep layers to shallow layers in FPN.


Despite the impressive results obtained by the FPN-based methods, they still suffer from \textbf{the inconsistency in gradient computation}\cite{liu2019learning}, thereby downgrading the effectiveness of FPN.
Since the anchors of deep layers can not match tiny objects, the corresponding positions on most deep layers are assigned as the negative sample, and only a few shallow layers assign them as the positive sample. 
Those corresponding features optimized toward the \textbf{negative sample} are directly delivered from deep layers to shallow layers by the element-wise addition operation (as shown in Fig. \ref{fig:challenge} (a)) to integrate with the features that are optimized toward the \textbf{positive sample}, which causes the inconsistency in gradient computation across different layers since the addition operation can not adaptively adjust the adjacent data flows. 
In other words, the gradient computation inconsistency downgrades deep layers' representation ability. Thereby, deep layers may not be able to guide the training of shallow layers but instead increase the burden on the shallow layers.

In this paper, we propose a Scale Selection Pyramid Network, namely SSPNet, to improve tiny person detection performance by mitigating the inconsistency in gradient computation across different layers.
We notice that the inconsistency in gradient computation does not occur in adjacent layers if an object is assigned as positive samples in adjacent layers, where the adjacent features corresponding to the object can be treated as suitable features of the adjacent layers since they are both optimized toward the positive sample.
This motivates us to design a Context Attention Module (CAM) to generate hierarchical attention heatmaps that point out objects of which scale can be assigned as positive samples in each layer of SSPNet. If those objects in adjacent attention heatmaps have intersections, the gradients in these intersected regions are consistent. 
Thus, we propose a Scale Selection Module (SSM) to deliver suitable features from deep layers to shallow layers with the guidance of heatmap intersections to resolve the inconsistency in gradient computation across different layers as shown in Fig. \ref{fig:challenge} (b). Besides, with the guidance of attention heatmaps, we also design a Scale Enhancement Module (SEM) to focus the detector on those objects of specific scales (assigned as positive samples) in each layer rather than vast and cluttered backgrounds.
Meanwhile, to further reduce false alarms, we employ a Weighted Negative Sampling (WNS) strategy to guide the detector to look more at representative samples to avoid missing representative samples in thousands of easy samples.

Our main contributions are summarized as follows:
\begin{itemize}
\item A novel SSPNet is proposed to suppress the inconsistency of gradient computation in FPN by controlling adjacent layers' data flow.

\item A WNS strategy is proposed to decrease false alarms by giving priority to those representative samples.

\item A mathematical explanation is given to explain why our SSPNet can relieve the inconsistency in gradient computation.

\item The proposed SSPNet significantly improves the detector's performance and outperforms SOTA detectors on the TinyPerson benchmark.

\end{itemize}

\begin{figure*}
	\centering
	\subfigure[]
	{
		\includegraphics[width=12cm,height=6cm]{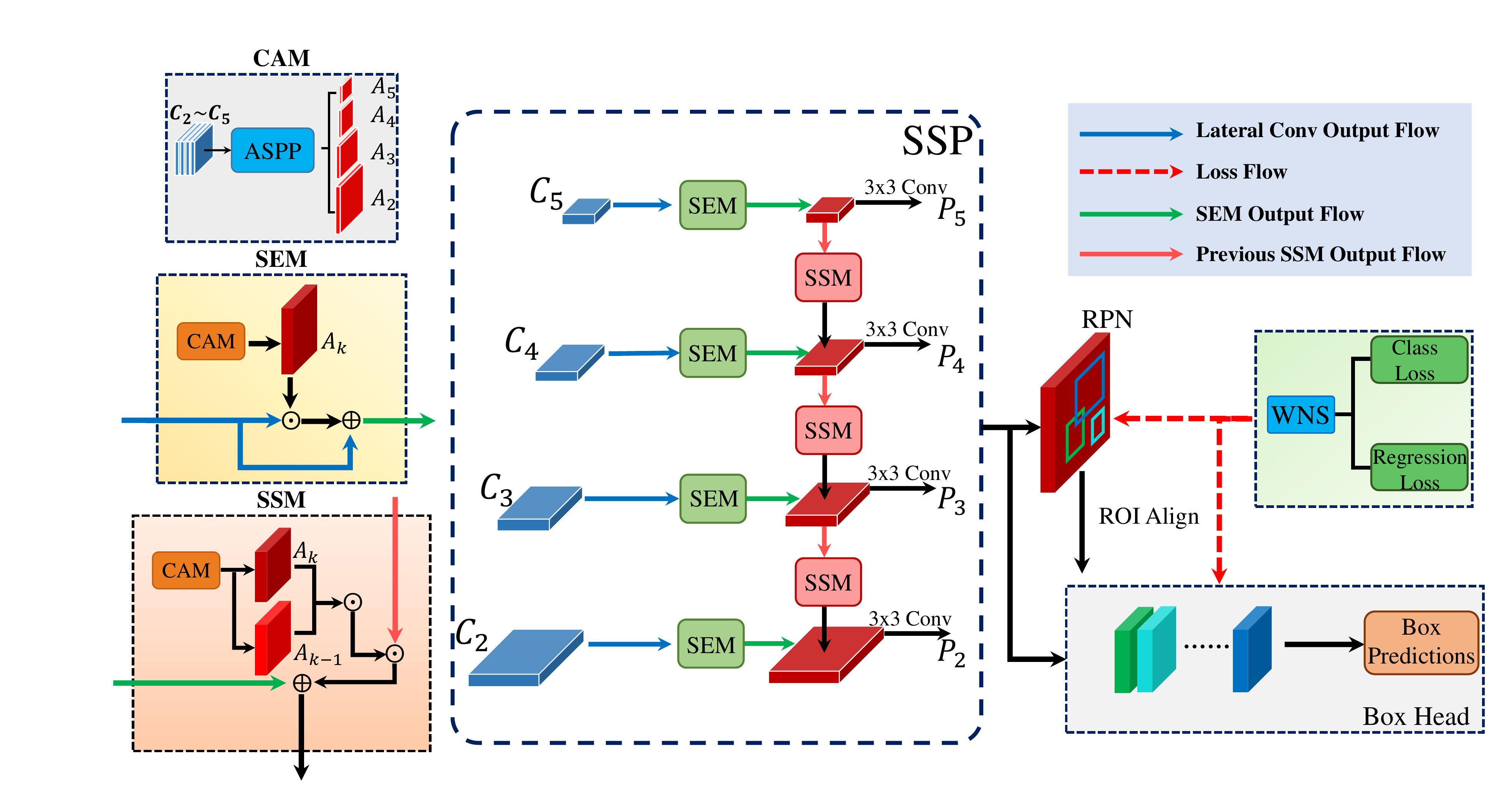}
	
	}
	\subfigure[]
	{
		\includegraphics[width=4cm,height=6cm]{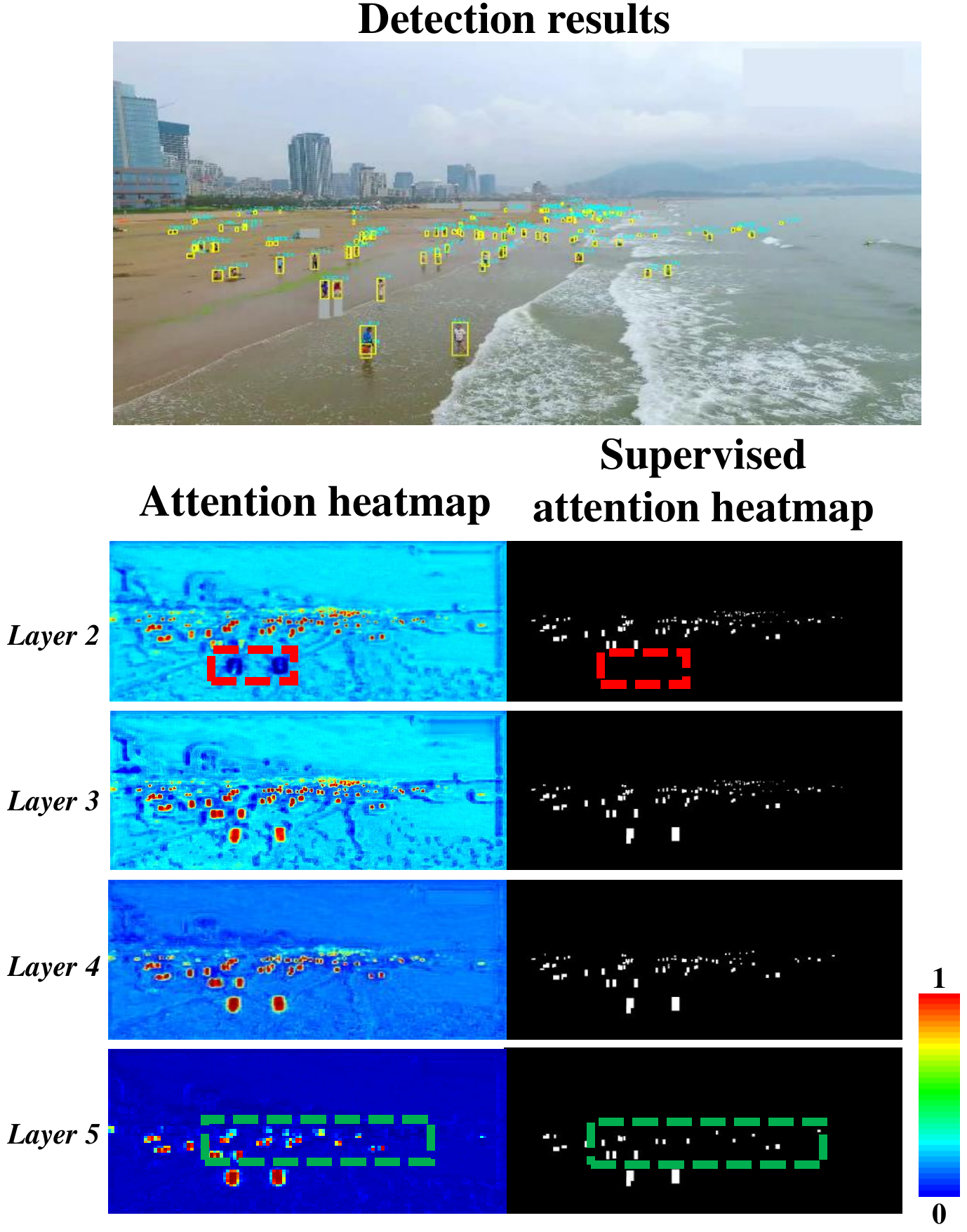}
		
	}
	\caption{(a) The overview of our proposed SSPNet. $P_{k}$ refers to the SSPNet's output at the $k^{th}$ layer. (b) Visualization of the attention heatmaps. The top row indicates the detection results, and the bottom row indicates the visualization of the attention heatmaps and the corresponding supervised attention heatmaps at different layers. The red dashed boxes contain large objects that are not objects of specific scales in the $1^{st}$ layer since they can not be matched by anchors of the $1^{st}$ layer, and the green boxes indicates most tiny objects that can not be matched by anchors of the $5^{th}$ layer.}
	\label{fig:architecture}
\end{figure*}

\section{Methodology}
Our SSPNet, based on the framework of Faster R-CNN\cite{ren2016faster}, includes CAM, SEM, SSM, as illustrated in Fig. \ref{fig:architecture} (a). In the following, we describe each module in detail.
\subsection{Context Attention Module}

To produce hierarchical attention heatmaps, we design CAM to generate attention heatmaps of different layers. As discussed in prior work\cite{yu2015multi,hu2017finding}, the context information can boost the performance of finding small objects. 
Thus, we first upsample all features produced by the backbone at different stages to the same shape as the bottom one and integrate them by concatenation. Then, atrous spatial pyramid pooling\cite{chen2017deeplab} (ASPP), with filters at multiple sampling rates and effective fields-of-views, is employed to find the object cues by considering multi-scale features. The context-aware features produced by ASPP are delivered to an activation gate that consists of multiple 3x3 convolutions with different strides and the sigmoid activation function to generate hierarchical attention heatmaps $A_{k}$:

\begin{equation}
	A_{k}=\sigma(\phi_{k}(F_{c},w,s)),
	\label{equ2}
\end{equation}
where $\sigma$ is the sigmoid activation function, $\phi_{k}$ denotes a 3x3 convolution at the $k^{th}$ layer, $w \in\mathbb{R}^{C_{F}\times1\times3\times3} $ represents the convolutional parameters, $F_{c}$ indicates the context-aware features produced by ASPP, and $s=2^{k-2}$ denotes the stride of  the convolution.

To point out which scale objects can be assigned as positive samples in each layer of SSPNet, we employ a supervised attention mechanism\cite{wang2017face} to highlight objects of specific scales in each layer of SSPNet and avoid being overwhelmed by vast backgrounds.
Specifically, the supervised attention heatmaps are associated with the objects matched by the anchors at different layers.
As shown in Fig. \ref{fig:architecture} (b), the supervised attention heatmap show different specific scale ranges at different layers, among which the red and green dashed boxes show that those objects that the corresponding layers' anchors do not match will be regarded as the background.
Moreover, the corresponding attention heatmaps are shown in Fig. \ref{fig:architecture} (b), and our CAM is able to yield the attention heatmaps with specific scale ranges.

\subsection{Scale Enhancement Module}
SEM is implemented to enhance the cues of objects of specific scales. Since the attention heatmaps at different layers have different scale preferences, allowing SEM to generate scale-aware features:
\begin{equation}
	F^{o}_{k} = (1+A_{k}) \odot F^{i}_{k},
	\label{equ3}
\end{equation}
where $F^{i}_{k}$ and $F^{o}_{k}$, respectively, are the input feature maps and output scale-aware features, $A_{k}$ is the attention heatmap at the $k^{th}$ layer, and $\odot$ is element-wise multiplication. Note that residual connection is used to avoid degrading the features around the objects since context information may facilitate detection.

\subsection{Scale Selection Module}

To select suitable features from deep layers for shallow layers, we propose the SSM to guide deep layers to deliver suitable features to shallow layers, where the suitable features do not cause inconsistency in gradient computation since they are optimized toward the same class. On the other hand, if the objects can be all detected in the adjacent layers, the deep layers will provide more semantic features and optimize with the next layer simultaneously\cite{kong2020foveabox}.
Our SSM can be formulated as follows:
\begin{equation}
	P'_{k-1} = (A_{k-1} \odot f_{nu}(A_{k}))\odot f_{nu}(P'_{k})+C_{k-1},
	\label{equ8}
\end{equation}
where the intersection of $A_{k-1}$ and $A_{k}$ is obtained by $\odot$, $f_{nu}$ denotes the nearest upsampling operation, $P'_{k}$ is the merged map at the ${k}^{th}$ layer, $C_{k-1}$ is the ${(k-1)}^{th}$ residual block’s output.

Specifically, SSM plays the role of scale selector. Those features corresponding to objects within the scale range of the next layer will be treated as suitable features to flow into the next layer, while other features will be weakened to suppress the inconsistency in gradient computation.

\subsection{Weighted Negative Sampling}
In large fields-of-view images captured by UAVs, complicated background typically introduces more noise than natural scene images. Besides, the partial occlusion in those images causes some objects to be annotated with only visible parts, resulting in the detector treats a person's parts as complete individuals, especially when the dataset is not large. Motivated by those considerations, we propose WNS to enhance the detector's generalization ability by looking more at representative samples.

Firstly, hard negative samples are usually regarded as positive ones with high confidence by the detector.
Thus, confidence is the most intuitive factor that needs to be considered. Then, to quantify the degree of incomplete objects, we adopt the intersection over foreground (IoF) \cite{yu2020scale} criterion.
Next, we construct a score-fusion function to consider the two factors of IoF and confidence as follows:
\begin{equation}
	s_{i} = \frac{e^{\lambda C_{i}+(1-\lambda)I_{i}}}{\sum_{i=1}^N e^{\lambda C_{i}+(1-\lambda)I_{i}}},
	\label{equ10}
\end{equation}
where $C_{i}$ and $I_{i}$, respectively, denote the $i^{th}$ detection result's confidence and the corresponding maximum IoF, and $\lambda$ indicates the balanced coefficient utilized to adjust the weights of IoF and confidence. Then, we can adjust the probability of selection for each sample based on $s_{i}$.

\subsection{Loss Function}
Our SSPNet can be optimized by a joint loss function, which is formulated as:
\begin{equation}
	L=L_{RPN}+L_{Head}+L_{A},
	\label{equ16}
\end{equation}
\begin{equation}
	L_{RPN} = \frac{1}{N_{bce}}\sum_{i=1}L_{bce}(rc_{i}, rc^{*}_{i})+\frac{\mu_{1}}{N_{reg}}\sum_{i=1}L_{reg}(rt_{i}, rt^{*}_{i}),
	\label{equ17}
\end{equation}
\begin{equation}
	L_{Head} = \frac{1}{N_{ce}}\sum_{i=1}L_{ce}(hc_{i}, hc^{*}_{i})+\frac{\mu_{2}}{N_{reg}}\sum_{i=1}L_{reg}(ht_{i}, ht^{*}_{i}),
	\label{equ18}
\end{equation}
where both $L_{RPN}$ and $L_{Head}$ employ a smooth L1 loss for bounding-box regression, but for classification, the former employs the binary cross-entropy (BCE) loss, while the latter employs cross-entropy loss. For $L_{RPN}$, $i$ is the index of bounding box in minibatch. $ rc_{i}$ and $rc^{*}_{i}$, respectively, denote the probability distributions of the predicted classes and the ground-truth. $rt_{i}$ and $ rt^{*}_{i}$, respectively, denote the predicted bounding box and ground-truth box. The classification and regression  losses are normalized by $N_{cls}$ (minibatch size) and $N_{reg}$ (number of boxes locations) and weighted by a balanced parameter $\mu_{1}$. By default we set $\mu_{1}$ and $\mu_{2}$ as 1. $L_{Head}$ is defined in a similar way.

$L_{A}$ indicates the attention loss to guide CAM to generate hierarchical attention heatmaps.
In particular, the attention loss can be formulated as:
\begin{equation}
	L_{A}=\alpha L^{b}_{A}+\beta L^{d}_{A},
	\label{equ12}
\end{equation}
where $\alpha$ and $\beta$, respectively, indicate the hyper-parameters of the dice loss\cite{milletari2016v} $L^{d}_{A}$ and the BCE loss $L^{b}_{A}$.
In detail, to avoid being overwhelmed by the vast backgrounds, we employ the dice loss to prioritize the foreground since it is only relevant to the intersection between the attention heatmap and the supervised attention heatmap.
Secondly, to remedy the gradient vanishing when the attention heatmap and the supervised attention heatmap have no intersection, we utilize BCE loss to deal with this extreme case and provide the valid gradient for optimizing. Moreover, we employ OHEM\cite{shrivastava2016training} to guarantee the detector focuses mainly on the non-object areas that are easily regarded as foreground, and we set the ratio of positives and negatives as 1:3 instead of considering all negatives. Specifically, the BCE loss is employed to learn poorly classified negatives, and the dice loss is employed to learn the class distribution to alleviate the imbalanced data.

\begin{table*}[t]
	\centering
	\caption{APs of different methods on TinyPerson.The best result in each MR is highlighted in bold, and the second best is highlighted in underline. $^{*}$: our implemented baseline.}
	\begin{tabular}{l|c|c|c|c|c|c|c|c}
		\toprule
		Detector        & $AP^{tiny}_{50} \uparrow $ & $AP^{tiny1}_{50} \uparrow $ & $AP^{tiny2}_{50}\uparrow $ & $AP^{tiny3}_{50}\uparrow $ & $AP^{small}_{50}\uparrow $ & $AP^{tiny}_{25}\uparrow $ & $AP^{tiny}_{75}\uparrow $ & $years$\\ \hline
		RetinaNet\cite{lin2017focal}                    &33.53   &12.24   &38.79   &47.38   &48.26    &61.51  &2.28   &2017\\ 
		Faster RCNN-FPN\cite{lin2017feature}              &47.35   &30.25   &51.58   &58.95   &63.18    &63.18  &5.83   &2017\\ 
		Faster RCNN-PANet$^{*}$(PAFPN)\cite{liu2018path}            &57.17   &44.53   &59.85   &65.52   &70.32    &77.30  &8.10   &2018\\ 
		Grid RCNN\cite{lu2019grid}                    &47.14   &30.65   &52.21   &57.21   &62.48    &68.89  &6.38   &2018\\ 
		Libra RCNN(Balanced FPN)\cite{pang2019libra}                   &44.68   &27.08   &49.27   &55.21   &62.65    &64.77  &6.26   &2019\\ 
		FCOS\cite{tian2019fcos}                         &17.90   &2.88    &12.95   &31.15   &40.54    &41.95  &1.50   &2019\\ 
		NAS-FPN\cite{ghiasi2019fpn}                      &37.75   &26.71   &40.69   &45.33   &52.63    &66.24  &3.10   &2019\\ 
		Faster RCNN-FPN-MSM\cite{yu2020scale}          &50.89   &33.79   &55.55   &61.29   &65.76    &71.28  &6.66   &2020\\ 
		Faster RCNN-FPN-MSM+\cite{jiang2021sm+}         &52.61   &34.20   &57.60   &63.61   &67.37    &72.54  &6.72   &2021\\
		RetinaNet+SM with S-$\alpha$\cite{Gong_2021_WACV} &52.56   &33.90   &58.00   &63.72   &65.69    &73.09  &6.64   &2021\\ 
		Swin-T\cite{liu2021swin}             &40.52   &31.92   &41.67   &47.06   &52.53    &59.42  &4.24  &2021\\ \hline
		
		RetinaNet$^{*}$ &52.86   &42.22   &58.07   &59.04   &66.40    &76.79  &6.5 &2017\\
		RetinaNet-SPPNet &54.66 ($\uparrow 1.8$)   &42.72   &60.16   &61.52   &65.24    &77.03  &6.31 &2021\\
		Cascade RCNN-FPN$^{*}$ &57.19   &45.21   &60.06   &65.06   &70.71    &76.99  &\underline{8.56} &2018\\
		Cascade RCNN-SSPNet &\underline{58.59} ($\uparrow 1.4$)   &\underline{45.75}   &\underline{62.03}   &\underline{65.83}   &\textbf{71.80}    &\underline{78.72}  &8.24 &2021\\
		Faster RCNN-FPN$^{*}$ &57.05   &43.82   &60.41   &65.06   &70.15 &76.39   &\underline{7.90} &2017\\
		Faster RCNN-SPPNet &\textbf{59.13} ($\uparrow$2.08)   &\textbf{47.56}  &\textbf{62.36}   &\textbf{66.15}   &\underline{71.17} &\textbf{79.47}   &\textbf{8.62}  &2021\\ \bottomrule
		
	\end{tabular}
	\label{tb:AP}
\end{table*}

\begin{table*}[t]
	\centering
	\caption{MRs of different methods on TinyPerson.The best result in each MR is highlighted in bold, and the second best is highlighted in underline. $^{*}$: our implemented baseline.}
	\begin{tabular}{l|c|c|c|c|c|c|c|c}
		\toprule
		Detector        & $MR^{tiny}_{50} \downarrow $ & $MR^{tiny1}_{50} \downarrow $ & $MR^{tiny2}_{50}\downarrow $ & $MR^{tiny3}_{50}\downarrow $ & $MR^{small}_{50}\downarrow $ & $MR^{tiny}_{25}\downarrow $ & $MR^{tiny}_{75}\downarrow $ & $years$\\ \hline
		RetinaNet\cite{lin2017focal}                      &88.31   &89.65   &81.03   &81.08   &74.05    &76.33  &98.76  &2017\\ 
		Faster RCNN-FPN\cite{lin2017feature}                &87.57   &87.86   &82.02   &78.78   &72.56    &76.59  &98.39  &2017\\ 
		Faster RCNN-PANet$^{*}$(PAFPN)\cite{liu2018path}           &85.18   &83.24   &77.39   &75.77   &65.38    &72.25  &98.32  &2018\\ 
		Grid RCNN\cite{lu2019grid}                     &87.96   &88.31   &82.79   &79.55   &73.16    &78.27  &98.21  &2018\\ 
		Libra RCNN (Balanced FPN)\cite{pang2019libra}                    &89.22   &90.93   &84.64   &81.62   &74.86    &82.44  &98.39  &2019\\ 
		FCOS\cite{tian2019fcos}                         &96.28   &99.23   &96.56   &91.67   &84.16    &90.34  &99.56  &2019\\ 
		NAS-FPN\cite{ghiasi2019fpn}                       &92.41   &90.37   &87.41   &87.50   &81.78    &77.79  &99.29   &2019\\ 
		Faster RCNN-FPN-MSM\cite{yu2020scale}           &85.86   &86.54   &79.20   &76.86   &68.76    &74.33  &98.23  &2020\\ 
		RetinaNet+SM with S-$\alpha$\cite{Gong_2021_WACV} &87.00   &87.62   &79.47   &77.39   &69.25    &74.85  &98.57  &2021\\ 
		Swin-T\cite{liu2021swin}           &89.91   &87.20   &85.44   &85.31   &80.28    &82.36  &98.89  &2021\\ \hline
		
		RetinaNet$^{*}$ &86.48   &82.40   &78.80   &78.18   &72.82    &70.52  &98.57 &2017\\
		RetinaNet-SPPNet$^{*}$ &85.30 ($\downarrow 1.18$)   &82.87   &76.73   &77.20   &72.37    &69.25  &98.63 &2021\\
		Cascade RCNN-FPN$^{*}$ &84.66   &\underline{82.32}   &76.84   &75.03   &64.77    &73.40  &\underline{98.18} &2018\\
		Cascade RCNN-SSPNet &\underline{83.47} ($\downarrow 1.19$)   &82.80   &\underline{75.02}   &\underline{73.52}   &\underline{62.06}    &\underline{68.93}  &98.27 &2021\\
		Faster RCNN-FPN$^{*}$ &84.12   &83.98   &76.10   &74.58   &64.03    &73.82  &98.19 &2017\\
		Faster RCNN-SPPNet &\textbf{82.79} ($\downarrow$1.33)   &\textbf{81.88}   &\textbf{73.93}   &\textbf{72.43}   &\textbf{61.26}   &\textbf{66.80} &\textbf{98.06}  &2021\\ \bottomrule
		
	\end{tabular}
	\label{tb:MR}
\end{table*}

\section{Experiments}

\subsection{Experimental Setup and Evaluation Metric}
We employ the TinyPerson benchmark\cite{yu2020scale} to evaluate our method’s effectiveness, and TinyPerson contains 794 and 816 images for training and inference, respectively. We crop each image into patches of $640\times 512$ pixels with $30$ pixels overlap in the training phase to avoid the GPU being out of memory. Our codes are based on the Mmdetection toolkit\cite{chen2019mmdetection}. We choose ResNet-50 as the backbone.
Motivated by YOLOv2\cite{Redmon_2017_CVPR}, we also utilize k-means clustering to find better prior anchors. By default, our SSPNet is trained in $10$ epochs, and the stochastic gradient descent is utilized as the optimizer with learning rate initialized as $0.002$ and decreased by a factor of $0.1$ after $8$ epochs. Besides, we set the proposal number of the RPN to $2000$ in the training phase and $ 1000$ in the testing phase, the loss weight $\alpha$ and $\beta$ are empirically set to $0.01$ and $1$, and the hyper-parameter $\lambda$ is set to $0.6$.

Keeping consistent with the TinyPerson benchmark\cite{yu2020scale}, we also adopt AP (average precision) as evaluation metrics. The TinyPerson benchmark\cite{yu2020scale} further divides $tiny[2, 20]$ into 3 sub-intervals: $tiny1[2, 8]$, $tiny2[8, 12]$, and $tiny3[12, 20]$.

\subsection{Comparison to State-of-the-arts}
We compare our SSPNet with some SOTA methods on TinyPerson.
As a standard criterion in the detection tasks, the higher the AP value, the better the detector’s performance. However, contrary to AP, as a criterion to reflect the percentage of objects being missed by the detector, the lower the MR value, the better the detector’s performance. 
Tables \ref{tb:AP} and \ref{tb:MR} show the performance of different methods in terms of MR and APmetrics. We found that most of the general-purpose detectors failed to achieve promising performance in such a scenario since there is a cross-domain gap between the UAV images and natural scene images.
Note that we do not directly adopt the performance of Yu et al.\cite{yu2020scale} as the baseline, but implement stronger baselines with data augmentation (such as ShiftScaleRotate, MotionBlur, etc.) and multi-scale training, which achieves better results than the SOTA detectors.
Note that all these detectors, i.e., RetinaNet, FasterRCNN, and Cascade RCNN, gain further improvements with our proposed SSPNet.

\subsection{Ablation Study}

\begin{table}[h]
	\caption{Effect of each component of Faster RCNN-SSPNet}
	\centering
	\begin{tabular}{cccccc}
		\toprule
		SEM & SSM & WNS &Attention Loss & $AP^{tiny}_{50}$            \\  \hline
		&                &                &                         &57.05               \\
		$\checkmark$   &                &                & $\checkmark$            &58.12 ($\uparrow1.07$)               \\
		& $\checkmark$   &                & $\checkmark$            &58.25 ($\uparrow1.20$)              \\
		&                & $\checkmark$   &                         &57.40 ($\uparrow0.35$)                \\
		$\checkmark$    & $\checkmark$   &                & $\checkmark$            &58.43 ($\uparrow1.38$)             \\
		$\checkmark$    & $\checkmark$   & $\checkmark$   &                         &57.86 ($\uparrow0.81$)              \\
		$\checkmark$    & $\checkmark$   & $\checkmark$   & $\checkmark$            &59.13 ($\uparrow2.08$)              \\  \bottomrule
	\end{tabular}
	\label{tb:componet}
\end{table}
\begin{figure*}[tbp]
	\centering
	\includegraphics[width=17cm,height=8cm]{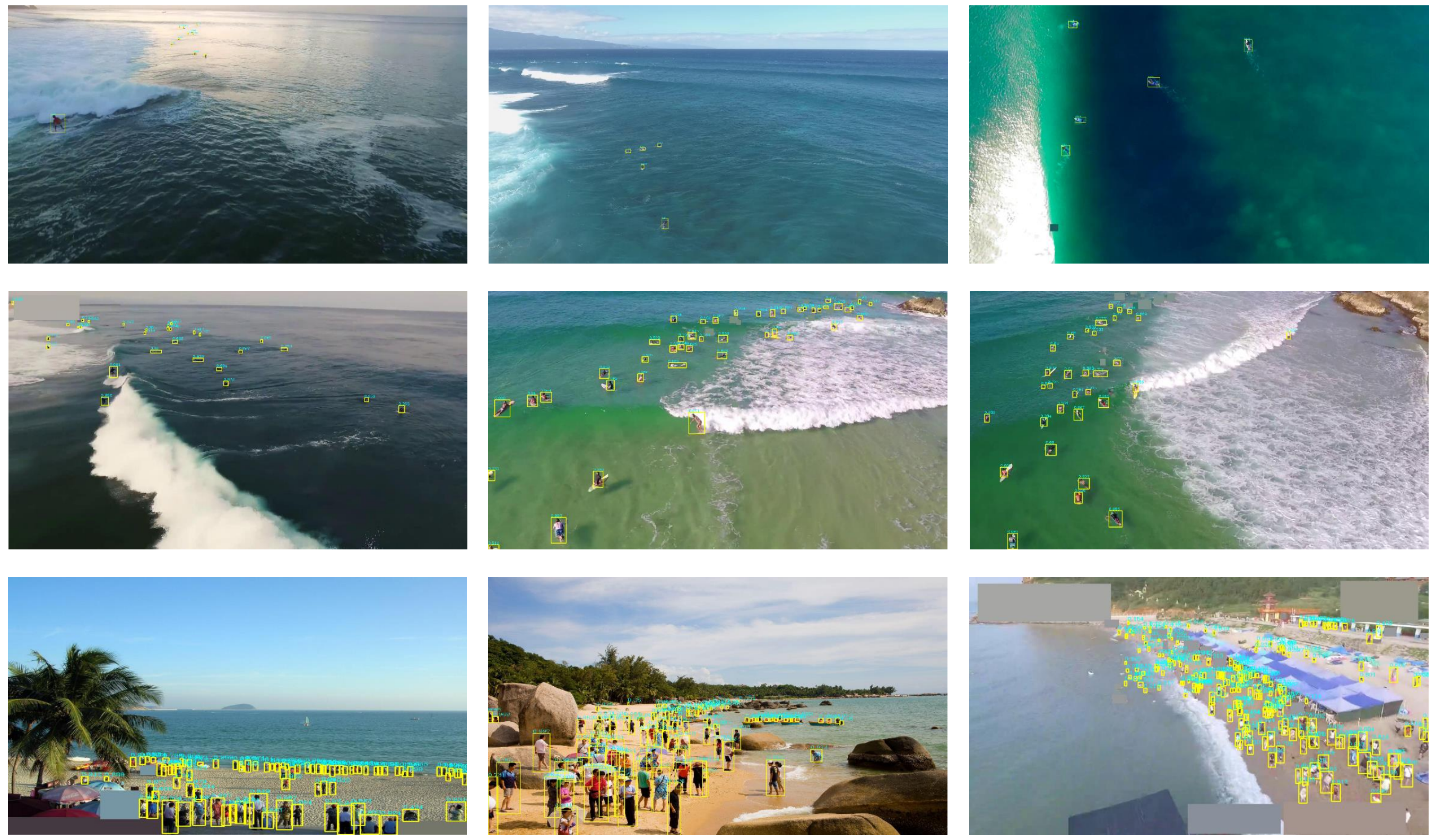}
	\caption{Visualization of detection results on TinyPerson’s test set, where those areas labeled as $uncertain$ in test set images are erased.}
	\label{fig:visual}
\end{figure*}
In this part, we analyze each module’s effect in our proposed method by applying the modules gradually. Results are reported in Table \ref{tb:componet}.
\begin{itemize}
\item \textbf{SEM}:
SEM is proposed to facilitate the detector to focus on objects of specific scales instead of the vast background. Results in Table \ref{tb:componet} prove its effectiveness with $1.07\%$ improvement in $AP^{tiny}_{50}$ over the baseline.
\item \textbf{SSM}:
SSM delivers those suitable features from deep layers to shallow layers and significantly improves
$AP^{tiny}_{50}$ of the baseline by $ 1.2\%$.
\item \textbf{WNS}:
WNS guides the detector to look more at those human parts and backgrounds that are easily considered as individuals. Without increasing more training parameters, WNS improves $AP^{tiny}_{50}$ of the baseline by $ 0.35\%$.
\item \textbf{Attention loss}:
In the last two rows of Table \ref{tb:componet}, we compare the attention loss with a single BCE loss, and the experimental results show that the attention loss can improve the performance by compensating for a single BCE loss’s shortcoming that only focuses on classification error at pixel-level.
\end{itemize}

\subsection{Influence of Balanced Coefficient}
\begin{figure}[h]
	\centering
	\includegraphics[scale=0.3]{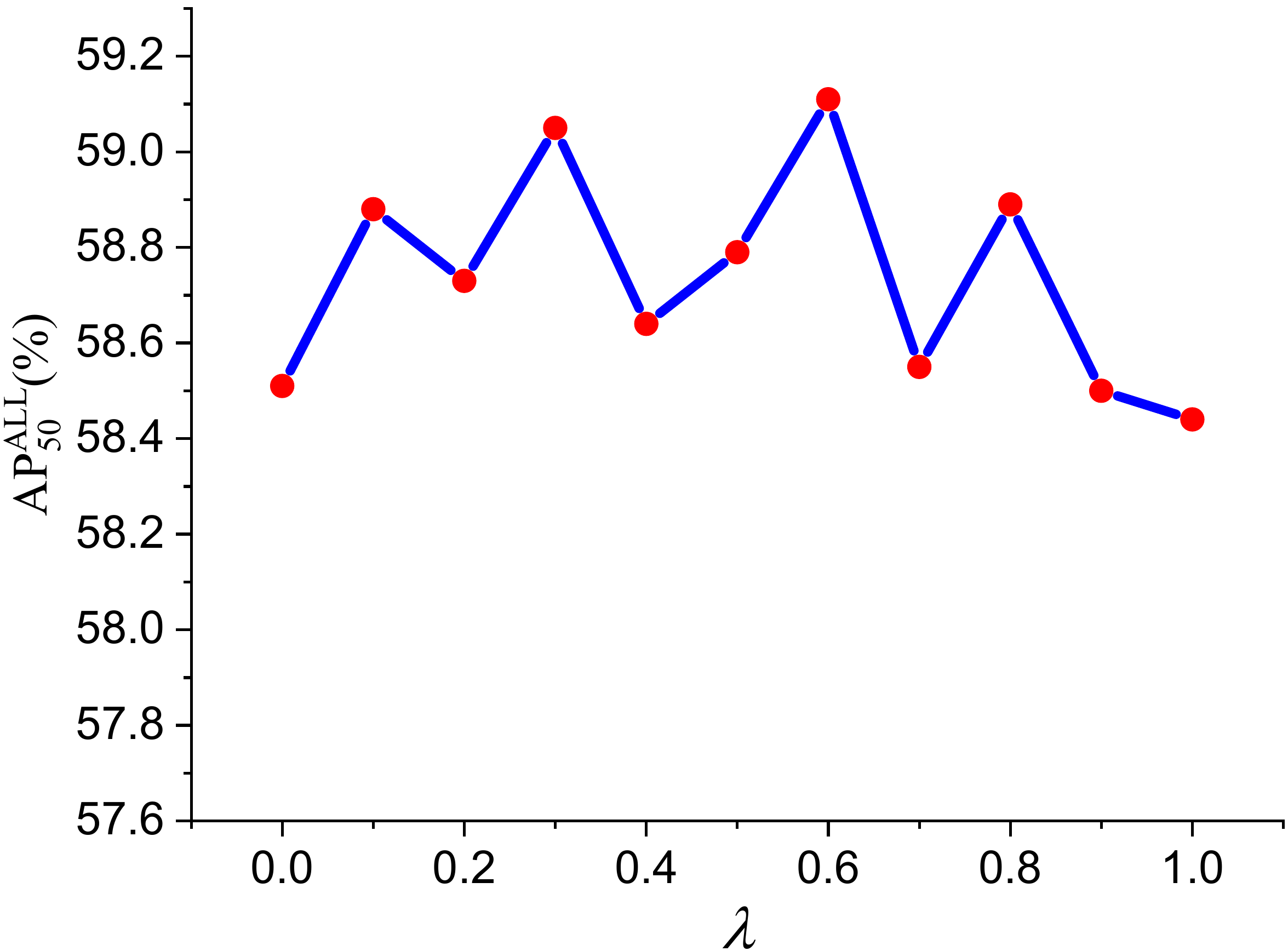}
	\caption{The impact of the balanced coefficient}
	\label{fig:coefficients}
\end{figure}
To choose an appropriate balanced coefficient, we investigate the impact of the balanced coefficient $\lambda$. We set $\lambda$ from 0.0 to 1.0, as shown in Fig. \ref{fig:coefficients}. When $\lambda$ is 0, it means that $s_{i}$ depends entirely on the   value of confidence and vice versa. Besides, we can observe from Fig. \ref{fig:coefficients}, the best performance is achieved when the coefficient is 0.6 while relying on only one of the two factors fails to achieve better performance. This might be because
most UAV images are accompanied by cluttered backgrounds or partial occlusion in the TinyPerson dataset. Therefore, we need to focus on those negative samples with both high confidence and high IoF.

\subsection{Effectiveness of Attention loss}
\begin{figure}[htbp]
	\centering
	\includegraphics[scale=0.4]{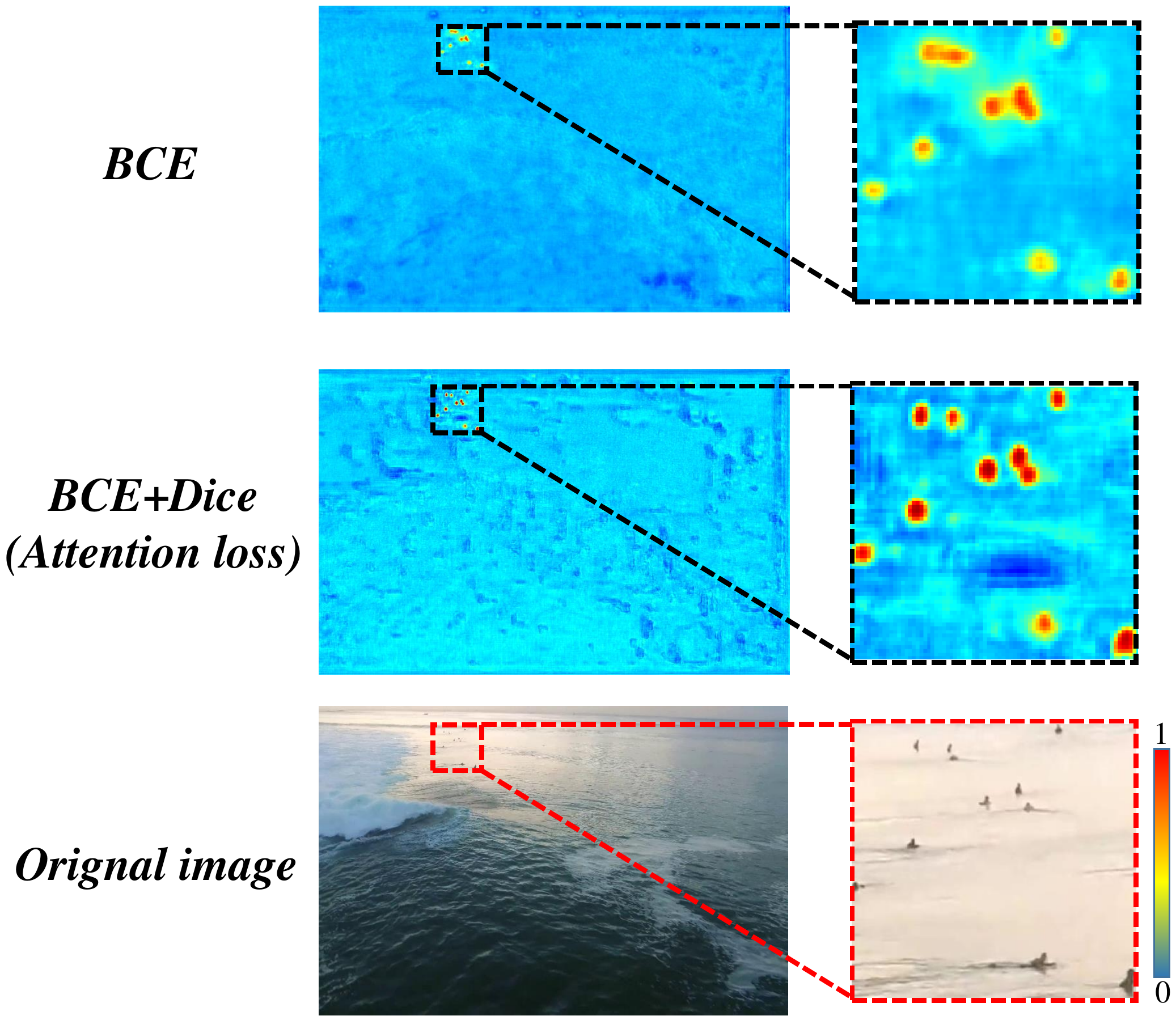}
	\caption{Illustration of the qualitative comparison on the attention loss.}
	\label{fig:hybrid loss}
\end{figure}
In this part, we present a qualitative evaluation to analyze why we employ the attention loss. As shown in Fig. \ref{fig:hybrid loss}, comparing the attention heatmap supervised by the attention loss with that supervised by a single BCE loss, the former attention map appears visually more apparent. On the contrary, the tiny objects' boundary in the attention map supervised by a single BCE loss is very blurred. This phenomenon can be attributed to that the dice loss is able to extract the foreground better.
Besides, as mentioned in Section II-E, since we only need to focus on suppressing those non-object areas that are easily regarded as foregrounds, the other non-objects areas will be retained adaptively to keep context information to facilitate detection. 
In other words, the attention heatmaps will not be entirely close to binary maps.

\subsection{Visual Analysis}

To further demonstrate the effectiveness of the proposed SSPNet, we show some visual detection results in Fig. \ref{fig:visual}, where the first two rows indicate the scenes of the persons in the sea, and the last row reflects the scenes of the persons on the land.
It can be observed from the first two rows that the postures of the persons on the surfboards have large variations, even some persons only reveal heads in backlight condition, but our SSPNet still can accurately recognize and locate them. Besides, the last row contains scale variations, crowd scenarios, and cluttered backgrounds. Even so, most of the persons can be recognized and located. Although the detector may miss a few persons due to occlusion, we believe that these problems can be significantly improved if the dataset can be more extensive and contains more of such crowded scenarios rather than being limited to a small number of training images.

\subsection{Mathematical Explanation}
In this part, we will analyze why our SSM can address the inconsistency in gradient computation from the perspective of gradient propagation. Without loss of generality, we discuss the gradient of a specific location $(i, j)$ in the $5^{th}$ layer. Suppose it is assigned as the center of a positive sample in the $2^{nd}$ layer and regarded as background in the other layers. The gradient $\frac{\partial L}{\partial P'_{5,(i,j)}}$  can be represented as follows:
\begin{equation}
	\begin{aligned}
		\frac{\partial L}{\partial P'_{5,(i,j)}}=\sum_{k=2}^{5}\frac{\partial L_{P_{k}}} {\partial P_{k,(i,j)}}  *\frac{\partial P_{k,(i,j)}}{\partial P'_{k,(i,j)}}*\frac{\partial P'_{k,(i,j)}}{\partial P'_{5,(i,j)}},
	\end{aligned}
	\label{equ:LP5}
\end{equation}
where $L_{P_{k}}$ indicates the loss of the $k^{th}$ layer in FPN. Since $P'_{k,(i,j)}$ is associated with $P'_{5,(i,j)}$ by an interpolation operation, $ \frac{\partial P'_{k,(i,j)}}{\partial P'_{5,(i,j)}}$ is usually a fixed value. In this case, $\frac{\partial L_{P^{2}}}{\partial P'_{2,(i,j)}}$will be optimized towards the positive sample, but $\frac{\partial L_{P_{3}}}{\partial P'_{3,(i,j)}}$, $\frac{\partial L_{P_{4}}}{\partial P'_{4,(i,j)}}$, and $\frac{\partial L_{P_{5}}} {\partial P'_{5,(i,j)}}$ will be optimized towards the negative sample, resulting in the inconsistency in gradient computation at the position $(i, j)$ in the $5^{th}$ layer.

For our SSM, we can express $P'_{k,(i,j)}$ as follows:
\begin{equation}
	\begin{aligned}
		P'_{k,(i,j)} = \underbrace{ A_{5,(i,j)}*A_{k,(i,j)}*\prod_{n=k+1}^{4}A^{2}_{n,(i,j)}}_{\psi_{k,(i,j)}}
		*P'_{5 \to k,(i,j)},
	\end{aligned}
\end{equation}
where $P'_{m \to l, (i,j)}$ indicates the merged map in SSPNet resized from layer $m$ to layer $l$ at the position $(i,j)$.

Next, we can adjust the inconsistency in gradient computation by the attention coefficient $\psi_{k,(i,j)}$ as follows:
\begin{equation}
	\begin{aligned}
		\frac{\partial P_{k,(i,j)}}{\partial P'_{5,(i,j)}} ={\psi_{k,(i,j)}}*
		\frac{\partial P_{k,(i,j)}}{\partial P'_{k,(i,j)}}*
		\frac{\partial P'_{5 \to k,(i,j)}}{\partial P'_{5,(i,j)}},
	\end{aligned}
\end{equation}
where the interpolation gradient $\frac{\partial P'_{m \to l,(i,j)}}{\partial P'_{m,(i,j)}}$ is usually a fixed value. Let us assume $\frac{\partial P'_{m \to l,(i,j)}}{\partial P'_{m,(i,j)}} \approx 1$. 
Then, we can simplify Equation (\ref{equ:LP5}) as:
\begin{equation}
	\begin{aligned}
		\frac{\partial L}{\partial P'_{5,(i,j)}} \approx \sum_{k=2}^{5}\frac{\partial L_{P_{k}}} {\partial P'_{k,(i,j)}} *\psi_{k,(i,j)}
	\end{aligned}
\end{equation}

\section{Conclusion}
This paper discusses the inconsistency of gradient computation encountered in FPN-based methods for tiny person detection, which weakens the representation ability of shallow layers in FPN.
We propose a novel model called SSPNet to overcome those challenges. Specifically, under the supervision of attention loss, CAM is able to yield the attention heatmaps with specific scale ranges at each layer in SSPNet. 
With the guidance of heatmaps, SEM strengthens those cues of objects of specific scales; SSM controls the data flow with intersection heatmaps to fulfill suitable feature sharing between deep layers and shallow layers.
Besides, to fully exploit small datasets to train better detectors, we propose a WNS to select representative samples to enable efficient training.
In future work, we will extend our method for a bidirectional feature pyramid network.

\ifCLASSOPTIONcaptionsoff
  \newpage
\fi



%

%
%
%
%

\bibliographystyle{ieeetran}
\bibliography{IEEEabrv,reference}

\begin{thebibliography}{10}
\providecommand{\url}[1]{#1}
\csname url@samestyle\endcsname
\providecommand{\newblock}{\relax}
\providecommand{\bibinfo}[2]{#2}
\providecommand{\BIBentrySTDinterwordspacing}{\spaceskip=0pt\relax}
\providecommand{\BIBentryALTinterwordstretchfactor}{4}
\providecommand{\BIBentryALTinterwordspacing}{\spaceskip=\fontdimen2\font plus
\BIBentryALTinterwordstretchfactor\fontdimen3\font minus
  \fontdimen4\font\relax}
\providecommand{\BIBforeignlanguage}[2]{{%
\expandafter\ifx\csname l@#1\endcsname\relax
\typeout{** WARNING: IEEEtran.bst: No hyphenation pattern has been}%
\typeout{** loaded for the language `#1'. Using the pattern for}%
\typeout{** the default language instead.}%
\else
\language=\csname l@#1\endcsname
\fi
#2}}
\providecommand{\BIBdecl}{\relax}
\BIBdecl

\bibitem{yu2020scale}
X.~Yu, Y.~Gong, N.~Jiang, Q.~Ye, and Z.~Han, ``Scale match for tiny person
  detection,'' in \emph{WACV}, 2020, pp. 1257--1265.

\bibitem{bai2018finding}
Y.~Bai, Y.~Zhang, M.~Ding, and B.~Ghanem, ``Finding tiny faces in the wild with
  generative adversarial network,'' in \emph{CVPR}, 2018, pp. 21--30.

\bibitem{noh2019better}
J.~Noh, W.~Bae, W.~Lee, J.~Seo, and G.~Kim, ``Better to follow, follow to be
  better: Towards precise supervision of feature super-resolution for small
  object detection,'' in \emph{CVPR}, 2019, pp. 9725--9734.

\bibitem{kisantal2019augmentation}
M.~Kisantal, Z.~Wojna, J.~Murawski, J.~Naruniec, and K.~Cho, ``Augmentation for
  small object detection,'' \emph{arXiv preprint arXiv:1902.07296}, 2019.

\bibitem{yu20201st}
X.~Yu, Z.~Han, Y.~Gong, N.~Jan, J.~Zhao, Q.~Ye, J.~Chen, Y.~Feng, B.~Zhang,
  X.~Wang \emph{et~al.}, ``The 1st tiny object detection challenge: Methods and
  results,'' \emph{arXiv preprint arXiv:2009.07506}, 2020.

\bibitem{liu2019learning}
S.~Liu, D.~Huang, and Y.~Wang, ``Learning spatial fusion for single-shot object
  detection,'' \emph{arXiv preprint arXiv:1911.09516}, 2019.

\bibitem{kong2020foveabox}
T.~Kong, F.~Sun, H.~Liu, Y.~Jiang, L.~Li, and J.~Shi, ``Foveabox: Beyound
  anchor-based object detection,'' \emph{TIP}, vol.~29, pp. 7389--7398, 2020.

\bibitem{Gong_2021_WACV}
Y.~Gong, X.~Yu, Y.~Ding, X.~Peng, J.~Zhao, and Z.~Han, ``Effective fusion
  factor in fpn for tiny object detection,'' in \emph{WACV}, January 2021, pp.
  1160--1168.

\bibitem{ren2016faster}
S.~Ren, K.~He, R.~Girshick, and J.~Sun, ``Faster r-cnn: Towards real-time
  object detection with region proposal networks,'' \emph{TPAMI}, vol.~39,
  no.~6, pp. 1137--1149, 2016.

\bibitem{yu2015multi}
F.~Yu and V.~Koltun, ``Multi-scale context aggregation by dilated
  convolutions,'' \emph{arXiv preprint arXiv:1511.07122}, 2015.

\bibitem{hu2017finding}
P.~Hu and D.~Ramanan, ``Finding tiny faces,'' in \emph{TPAMI}, 2017, pp.
  951--959.

\bibitem{chen2017deeplab}
L.-C. Chen, G.~Papandreou, I.~Kokkinos, K.~Murphy, and A.~L. Yuille, ``Deeplab:
  Semantic image segmentation with deep convolutional nets, atrous convolution,
  and fully connected crfs,'' \emph{TPAMI}, vol.~40, no.~4, pp. 834--848, 2017.

\bibitem{wang2017face}
J.~Wang, Y.~Yuan, and G.~Yu, ``Face attention network: An effective face
  detector for the occluded faces,'' \emph{arXiv preprint arXiv:1711.07246},
  2017.

\bibitem{milletari2016v}
F.~Milletari, N.~Navab, and S.-A. Ahmadi, ``V-net: Fully convolutional neural
  networks for volumetric medical image segmentation,'' in \emph{3DV}.\hskip
  1em plus 0.5em minus 0.4em\relax IEEE, 2016, pp. 565--571.

\bibitem{shrivastava2016training}
A.~Shrivastava, A.~Gupta, and R.~Girshick, ``Training region-based object
  detectors with online hard example mining,'' in \emph{CVPR}, 2016, pp.
  761--769.

\bibitem{lin2017focal}
T.-Y. Lin, P.~Goyal, R.~Girshick, K.~He, and P.~Doll{\'a}r, ``Focal loss for
  dense object detection,'' in \emph{ICCV}, 2017, pp. 2980--2988.

\bibitem{lin2017feature}
T.-Y. Lin, P.~Doll{\'a}r, R.~Girshick, K.~He, B.~Hariharan, and S.~Belongie,
  ``Feature pyramid networks for object detection,'' in \emph{CVPR}, 2017, pp.
  2117--2125.

\bibitem{lu2019grid}
X.~Lu, B.~Li, Y.~Yue, Q.~Li, and J.~Yan, ``Grid r-cnn,'' in \emph{CVPR}, 2019,
  pp. 7363--7372.

\bibitem{tian2019fcos}
Z.~Tian, C.~Shen, H.~Chen, and T.~He, ``Fcos: Fully convolutional one-stage
  object detection,'' in \emph{ICCV}, 2019, pp. 9627--9636.

\bibitem{ghiasi2019fpn}
G.~Ghiasi, T.-Y. Lin, and Q.~V. Le, ``Nas-fpn: Learning scalable feature
  pyramid architecture for object detection,'' in \emph{CVPR}, 2019, pp.
  7036--7045.

\bibitem{pang2019libra}
J.~Pang, K.~Chen, J.~Shi, H.~Feng, W.~Ouyang, and D.~Lin, ``Libra r-cnn:
  Towards balanced learning for object detection,'' in \emph{CVPR}, 2019, pp.
  821--830.

\bibitem{liu2018path}
S.~Liu, L.~Qi, H.~Qin, J.~Shi, and J.~Jia, ``Path aggregation network for
  instance segmentation,'' in \emph{CVPR}, 2018, pp. 8759--8768.

\bibitem{jiang2021sm+}
N.~Jiang, X.~Yu, X.~Peng, Y.~Gong, and Z.~Han, ``Sm+: Refined scale match for
  tiny person detection,'' in \emph{ICASSP}.\hskip 1em plus 0.5em minus
  0.4em\relax IEEE, 2021, pp. 1815--1819.

\bibitem{liu2021swin}
Z.~Liu, Y.~Lin, Y.~Cao, H.~Hu, Y.~Wei, Z.~Zhang, S.~Lin, and B.~Guo, ``Swin
  transformer: Hierarchical vision transformer using shifted windows,''
  \emph{arXiv preprint arXiv:2103.14030}, 2021.

\bibitem{chen2019mmdetection}
K.~Chen, J.~Wang, J.~Pang, Y.~Cao, Y.~Xiong, X.~Li, S.~Sun, W.~Feng, Z.~Liu,
  J.~Xu \emph{et~al.}, ``Mmdetection: Open mmlab detection toolbox and
  benchmark,'' \emph{arXiv preprint arXiv:1906.07155}, 2019.

\bibitem{Redmon_2017_CVPR}
J.~Redmon and A.~Farhadi, ``Yolo9000: Better, faster, stronger,'' in
  \emph{CVPR}, July 2017.

\end{thebibliography}

%




\end{document}